\documentclass[letterpaper, 10 pt, conference]{ieeeconf}  

\IEEEoverridecommandlockouts                              

\usepackage{graphics} 
\usepackage{epsfig} 
\usepackage{mathptmx} 
\usepackage{times} 
\usepackage{amsmath, bm} 
\usepackage{amssymb}  
\usepackage{amsbsy}
\usepackage{algorithm}
\usepackage{color}
\usepackage{algpseudocode}
\usepackage{tabularx}
\usepackage{multirow}
\usepackage{booktabs}
\usepackage{caption}
\usepackage{soul}
\newcommand\norm[1]{\left\lVert#1\right\rVert}

\title{\LARGE \bf
Intention Communication and Hypothesis Likelihood \\ in Game-Theoretic Motion Planning
}

\author{Makram Chahine$^{1}$, Roya Firoozi$^{2}$, Wei Xiao$^{1}$, Mac Schwager$^{2}$ and Daniela Rus$^{1}$
\thanks{$^{1}$ Computer Science and Artificial Intelligence Lab, Massachusetts Institute of Technology \texttt{{\small \{chahine, weixy, rus\}@mit.edu}}}
\thanks{$^{2}$ Department of Aeronautics \& Astronautics, Stanford University \texttt{{\small \{rfiroozi, schwager\}@stanford.edu}}}
}

\begin{document}

\maketitle
\thispagestyle{empty}
\pagestyle{empty}

\begin{abstract}


Game-theoretic motion planners are a potent solution for controlling systems of multiple highly interactive robots. Most existing game-theoretic planners unrealistically assume \textit{a priori} objective function knowledge is available to all agents. 
To address this, we propose a fault-tolerant receding horizon game-theoretic motion planner that leverages inter-agent communication with intention hypothesis likelihood. Specifically, robots communicate their objective function incorporating their intentions. A discrete Bayesian filter is designed to infer the objectives in real-time based on the discrepancy between observed trajectories and the ones from communicated intentions. In simulation, we consider three safety-critical autonomous driving scenarios of overtaking, lane-merging and intersection crossing, to demonstrate our planner's ability to capitalize on alternative intention hypotheses to generate safe trajectories in the presence of faulty transmissions in the communication network.

\end{abstract}

\section{Introduction}

Robot control in interactive environments is very challenging due to the complexity of evaluating the impact of an agent's actions on the behaviour of others. Predict-then-plan strategies have substantially been addressed in the literature; trajectories of other agents are predicted first, then used as constraints in a single-agent planning scheme. However this decoupling ignores the inherently interactive nature of the problem. Indeed, in order to capture the reactive nature of agents in a scene, a robot must be able to simultaneously predict the trajectories of other agents while planning its own trajectory. Differential game theory provides a suitable framework for expressing these types of multi-agent planning problems without requiring \textit{a priori} predictive assumptions. Solvers exist to find Nash equilibrium trajectories for such problems in real-time, and although they provide open-loop solutions to the motion planning problem over the time horizon considered, repeatedly resolving the game as new information is processed generates a policy that is closed-loop in the model-predictive control (MPC) sense.

The main challenge of decentralized implementation of game-theoretic motion planners is that common knowledge of the dynamics, constraints and objective functions of all partaking players is required. One can argue that dynamics and constraints could be assumed or inferred in structured environments e.g. in driving scenarios. However, for a player to assign an objective function to each of the other agents, they must have a representation of their intentions.  We propose to use vehicle-to-vehicle (V2V) communication for vehicles to share their intentions with one another. 

Robotics settings can hugely benefit from communication technologies.
In the current robotics settings, i.e. autonomous driving, each agent acts as an isolated agent, choosing its own control decisions solely based on its own on-board sensing information. 
\begin{figure}[t]
	\centering
	\epsfig{file=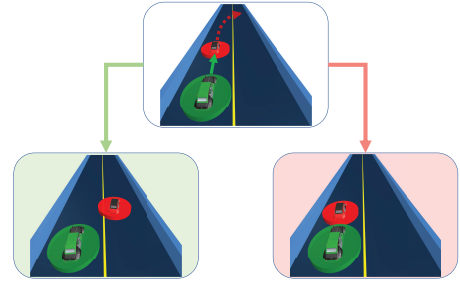, width=0.37\textwidth} 
	\caption{Overtaking on a highway. As the red car signals to the right, the green car anticipates it making way and accelerates. If the red car's signalling is reliable and it does move over, the green car would have capitalized on this communicated intention to reach its objective faster (left scenario). On the other hand, if the red car does not switch lanes, the green car runs the risk of collision if it is unable to react in time (right scenario).}
	\label{fig:teaser}
    $\vspace{-8mm}$
\end{figure}
However, the collaborative communication of agents' intentions over a V2V network can significantly improve the quality of planned trajectories, as vehicles can reason about the actions that others might take to reach their objective. A simple and very frequent real-life example is that of overtaking, where knowledge of the other vehicle's intention allows for a more efficient maneuver as shown in Fig. \ref{fig:teaser}. In the game-theoretic framework, the communication of intention is best formalized as the sharing of one's cost function, which directly intervenes in computing the Nash equilibrium trajectory solution. In cooperative settings, we could thus design collaborative planning solutions requiring agents to communicate their cost functions at regular intervals. In turn, this raises questions about the security and robustness of the communication protocol to breaches and faults in safety-critical applications. In the example shown above in Fig. \ref{fig:teaser}, we present a case where faulty communication could lead to a collision.

In this paper, we investigate the use of V2V communication as a solution to the safety shortcomings arising from cost function mismatch in MPC implementations of game-theoretic motion planners. We specify the requirements such a solution would need to satisfy, namely robustness to communication faults and adaptability throughout the unfolding of an evolving multi-agent scene. The following assumptions are made in our proposed framework:
\begin{enumerate}
    \item A cooperative setting is assumed in which vehicles are connected using a V2V communication network.
    \item A set of possible intentions (hypotheses) for each vehicle 
    is available to be used in real time.    
\end{enumerate}
Our proposed planner updates hypothesis likelihoods, allowing an agent to reassess the reliability of communicated intentions coming from each player in the scene. With a hypothesis likelihood assigned to each of the other players intentions, the agent adapts its version of a dynamic game at each replanning interval. Our main contributions are,
\begin{enumerate}
    \item An online Bayesian inference approach to estimate the relative likelihoods of available hypotheses based on past observations in the context of dynamic games with communicated cost functions,
    \item A communication-based game-theoretic receding horizon motion planner is designed to be resilient to faulty transmissions and safer than naive MPC implementations of dynamic game motion planners,
    \item Simulations on three autonomous driving scenarios: overtaking, lane-merging and intersection crossing, empirically showcasing the superiority of our approach in planning safe trajectories and avoiding crashes.
\end{enumerate}


\section{Related Work}

\textbf{Dynamic Games.} Differential game theory provides a framework to model multi-agent interactions without requiring a priori assumption about predicting other agents' behaviors \cite{Starr1969}. Differential games are characterized by solving Hamilton-Jacobi equations in which all agent’s Hamiltonians are coupled with one another. However, most of these methods do not admit analytical solutions, suffer from the curse of dimensionality and cannot be solved efficiently using numerical techniques. To make the problem tractable, iterative linear-quadratic (iLQ) games are proposed in \cite{David2020},\cite{Schwarting2021}. In the iLQ games at each iteration, the problem is solved locally by successive approximation using linear dynamics and quadratic cost. The iterative linear quadratic game can be solved as a coupled Riccati differential equation and provides feedback strategies which form a feedback Nash equilibrium of the game \cite{Li1995}, \cite{Mukai2001}. These approximation techniques are computationally efficient and suitable for real-time applications, but since the feedback policies are fixed, these approaches cannot adapt to any change in the behavior of the other agents. In other words, the game structure is assumed to remain fixed. In contrast, our proposed approach allows adaptability to any change in the game structure and provides open-loop Nash equilibrium, instead. 

\textbf{Game-theoretic planning} A game-theoretic setting allows to model both adversarial and cooperative agents. In fact, scenarios such as autonomous driving, where robots need to interact with other intelligent agents, are fundamentally game-theoretic \cite{Fisac2018, Trautman2010}, \cite{Dreves2018}. Formalizing such interactions as a game, robots can weigh the impact of their decisions on the actions of other agents \cite{Sadigh2016}, even in competitive settings such as drone or car racing \cite{Spica2018,Wang2021}. Most game-theoretic works do not assume availability of a communication network among the agents and model the interactions as a game formulation. In \cite{peters2021inferring, laine2021multi, le2021lucidgames} data-driven methods are presented to infer objective functions of other agents online in game theoretic planning without communication. On the other hand distributed optimization approaches assume the existence of a communication network and solve the optimization based on the received communicated predictions \cite{Felix2018}, \cite{Ferranti2018}. They devise a communication and planning protocol in which all the agents share their predicted planned trajectories. One drawback of the former approach is that in an uncertain and dynamic environment such as autonomous driving setting, solving a game with a fixed structure may not be robust with respect to adversarial agents' behavior. Also, distributed optimization approaches that completely rely on the communicated trajectories and make use of predictions in the planning are susceptible to adversarial cyber-attacks. In contrast, our proposed approach considers a combination of the two, in which we consider solving a game while a communication network is assumed among the agents and they make use of communicated information to update their reliability towards other agents. Within the receding-horizon game-theoretic framework the agents update the game cost function according to the communicated information about other agents' intentions. Incorporating reliability into the game formulation leads to quick adaptation and promotes resilience against failures. 

\textbf{Communication-based planning and control.}  Based on vehicle-to-infrastructure (V2I) communication, the
emergence of connected and automated vehicles \cite{Schrank2015}, \cite{Milanes2012}, \cite{Youssef2019}
has the potential to drastically improve the performance of vehicles in traffic bottlenecks (such as merging, intersection, etc.), ultimately
saving energy, and avoiding congestion and accidents. Optimal control problem are used in some of these
approaches \cite{Xiao2021}, while Model Predictive Control (MPC) techniques are employed as an alternative \cite{Ntousakis2016}, mainly to take additional constraints and disturbances into account \cite{Mohamad2018}. All the above mentioned works assume that information shared through V2I is trust-worthy. While in this work, we consider communication-based control in the presence of faulty transmissions on the communication network.

\section{Problem Statement}


We consider a sequence of Generalized Nash Equilibrium Problems (GNEPs) involving $N$ players $i \in \{1,\dots,N\}$ over a horizon of $H$ time steps. An agent $i$'s state at time step index $t$ is denoted $\mathbf{x}_t^i \in \mathbb{R}^{n^i}$ and control input $\mathbf{u}_t^i \in \mathbb{R}^{m^i}$, with dimensions of agent $i$'s state and control $n^i$ and $m^i$. Let $\mathbf{x}_t = [\mathbf{x}_t^{1, \top}, \dots, \mathbf{x}_t^{N, \top}]^\top \in \mathbb{R}^{n}$ denote the joint state and $\mathbf{u}_t = [\mathbf{u}_t^{1, \top}, \dots, \mathbf{u}_t^{N, \top}]^\top \in \mathbb{R}^{m}$ denote the joint control of all agents at time $t$, with joint dimensions $n = \sum_i n^{i}$ and $m = \sum_i m^i$. We define player $i$'s policy as $\pi^i = [\mathbf{u}_1^{i, \top}, \dots, \mathbf{u}_{H-1}^{i, \top}]^\top \in \mathbb{R}^{\Tilde{m}^i}$  where $\Tilde{m}^i = m^i (H-1)$ denotes the dimension of the entire trajectory of agent $i$'s control inputs. The notation $\neg i$ indicates all agents except $i$, for instance $\pi^{\neg i}$ represents the vector of the agents' policies except that of $i$. Also, let $X = [\mathbf{x}_1^{\top}, \dots, \mathbf{x}_{H-1}^{\top}]^\top \in \mathbb{R}^{\Tilde{n}}$, with $\Tilde{n}= n(H-1)$, denote the trajectory of joint state variables resulting from the application of the joint control inputs to the dynamical system defined by $f:\mathbb{R}^{n} \times \mathbb{R}^{m} \rightarrow \mathbb{R}^{n}$ such that,
\begin{align}
    \mathbf{x}_{t+1} = f(\mathbf{x}_{t}, \mathbf{u}_{t}) \label{dyna}
\end{align}
Over the whole trajectory we can express the above kinodynamic constraints with $\Tilde{n}$ equality constraints,
\begin{align}
    D(X,\pi^1,\dots,\pi^N) = D(X,\pi) = 0 \in \mathbb{R}^{\Tilde{n}} \label{kino}
\end{align}
The cost function of each player $i$ depends on its policy $\pi^i$ as well as on the joint state trajectory $X$, which is common to all players, such that $\forall i \in \{1,\dots,N\}$,
\begin{align}
    J^i(X,\pi^i) = c_H^i(\mathbf{x}_H) + \sum_{t=1}^{H-1} c_t^i(\mathbf{x}_t, \mathbf{u}_t^{i}) \label{cost}
\end{align}
Notice that as player $i$ minimizes $J^i$ with respect to $X$ and $\pi^i$, the selection of $X$ is constrained by the other
players’ strategies $\pi^{\neg i}$ and the dynamics of the joint system via (\ref{kino}). In addition, the strategy $\pi^i$ could be required to satisfy (safety) constraints that depend on the joint state trajectory $X$ as well as on the other players strategies $\pi^{\neg i}$ . This can be expressed with a set of $g$ inequality constraints,
\begin{align}
    C(X,\pi) \leq 0 \in \mathbb{R}^{g} \label{cons}
\end{align}
where $C:\mathbb{R}^{\Tilde{n}} \times \mathbb{R}^{\Tilde{m}}\rightarrow \mathbb{R}^g$.
The GNEP we form is the problem of minimizing (\ref{cost}) for all players $i \in \{1,\dots,N\}$ with respect to (\ref{kino}) and (\ref{cons}). More specifically,
\begin{align}
    \underset{X,\pi^i}{\min}\ & J^i(X,\pi^i) \qquad \forall i \in \{1,\dots,N\} \nonumber\\
    \text{subject to }\ & D(X,\pi) = 0 \label{gnep}\\
    & C(X,\pi) \leq 0 \nonumber
\end{align}
The solution to such a dynamic game is a generalized Nash equilibrium, i.e. a policy $\pi$ such that, $\forall i \in \{1,\dots,N\}$, $\pi^i$ is a solution to (\ref{gnep}) with the other players' policies given by $\pi^{\neg i}$ that is also solved by (\ref{gnep}) for all $\neg i$. As a consequence, at a Nash equilibrium point solution, no player can improve their strategy by unilaterally modifying their policy.

\subsection*{MPC implementation with cost updates}
Cost structures are fixed during the resolution of a dynamic game, which can lead to dangerous trajectories when costs used differ from the true expressions. Solutions with state feedback such as iLQG, or MPC implementations of open-loop solvers have been shown to mitigate this issue.
However, these solutions do not tackle the root of the problem which is the assumed versus true cost mismatch. Furthermore, in real-life scenarios, a robot's representation of its environment evolves through sensor measurement and inter-agent communication. Hence, intelligent agents should be able to update their interpretation of others' intentions. In our game-theoretic framework, this is equivalent to enabling objective function estimate updates at relevant frequencies. Proposed methods to achieve this include inverse optimal control algorithms and Inverse Reinforcement Learning (IRL).

To the best of our knowledge, this problem has not yet been tackled for intention-communicating agents with possible transmission faults corrupting the communication channel. 
We design a communication-based game-theoretic receding horizon motion planner to update the cost representations agents have of others by capitalizing on truthful communication and absorbing misleading transmissions to avoid dangerous trajectories.


\section{Hypothesis Adaptive Motion Planner}

Our solution relies on updating the perceived costs by comparing executed trajectories to those obtained by solving the set of games corresponding to various intention hypotheses. This approach is based upon the observation that the set of broadly defined possible intentions in many multi-robot motion planning is small, and that solutions that are close most likely arise from cost formulations that are also close. The latter assumption might not hold for arbitrary systems with complex nonlinear dynamics, however is sound for agents only interacting via collision constraints. The non-uniqueness of Nash equilibria also means that even with correct costs, potentially large trajectory deviations can arise. However, Peters et al. \cite{peters2020align} show that it is possible to ensure agents agree on a Nash equilibrium.

We denote agent $i$'s true cost function as $c_v^i$. Agent $i$ broadcasts a communicated cost denoted $c_c^i$ to all other agents in its vicinity. (In the most general setting, an agent could communicate different cost functions to different agents in the scene, and could modify the communicated cost at each MPC iteration). We also have access in a database to a set of $\xi$ possible hypotheses for other agents intentions,
notably for agent $i$, cost functions $c_{h,I}^i$ for $I \in \{1,2,\dots, \xi\}$.
We denote $c^i_h$ the vector of hypothetical cost functions agent $i$ has about the scene such that $c^i_h = [c_{h,1}^i, \dots ,c_{h,\xi}^i]$.
Agent $i$ solves a recursive game and must estimate $\xi+1$ probability parameters representing the relative likelihood of each of the hypotheses, $\lambda_I$ with $I \in \{0, \dots, \xi\}$ and such that $\sum_{I=0}^ \xi \lambda_I = 1$ and $\lambda_I\geq 0$, then uses the estimated likelihoods to assign a cost for agent $j$ denoted $c_j^i$.
Omitting the $i$, the ego vehicle solves a non-cooperative game with its own cost $c_v$ and the assigned cost function $c_j$ for the ado agent $j$ according to,
\begin{align}
    k_{\textrm{max}} = \mathrm{argmax}([\lambda_0, \dots, \lambda_{\xi}]) \label{kay}\\
    c_j = \begin{cases} \mbox{$c_c^j$\quad,} & \mbox{if } k_{\textrm{max}}=0 \\ \mbox{$c_{h,k_{\textrm{max}}}$\quad,} & \mbox{otherwise} \end{cases} \label{cjay}
\end{align}

The design choice of planning with the highest likely hypothesis is motivated by the discrete nature of possible cost functions in most driving scenarios (switching lanes, turning left or right etc.). In games with continuous cost structure (desired speed, aggressiveness, comfort metrics), we propose using the weighted sum of the intention hypotheses to determine the cost assigned to agent $j$,
\begin{align}
    c_j =  \lambda_0 c_c^j + \sum_{I=1}^\xi \lambda_I c_{h,I} \label{cjay_prime}
\end{align}


\subsection{Online Bayesian inference for hypothesis estimation}\label{filter}

We use online discrete Bayesian filtering to update the distribution weights. The belief vector $\boldsymbol{\Lambda}$ is defined as $\boldsymbol{\Lambda} = [\lambda_0, \lambda_1, ...\lambda_{\xi}]^T$. The transition matrix is defined as $\mathbf{T} = [T{_{k l}}]_{k,l= 0}^\xi$, where each entry of the matrix is $T{_{k l}} = p(I_{t+1} = k| I_t = l)$. The filter predict step is, 
\begin{align}\label{predict_step}
    \boldsymbol{\Lambda}_{t+1|t} = \mathbf{T} \boldsymbol{\Lambda}_{t|t}
\end{align}
where the subscript $t|t'$ denotes the estimate of $\boldsymbol{\Lambda}$ at time $t$ given the observations up to and including time $t'$. 
To make minimal assumption on how the intention dynamics evolve, we take $\mathbf{T}$ to be the identity matrix. 
The measurement matrix $\mathbf{M}$ is defined as,
\begin{align}
    \mathbf{M} = \textrm{diag}[p(\mathbf{y}_{t-s:t}|I_t = 0),...,p(\mathbf{y}_{t-s:t}|I_t = \xi)] 
\end{align}
where the measurement likelihood for diagonal elements of the matrix $\mathbf{M}$ are computed based on the steps outlined in Algorithm \ref{alg:filtering}. The filtering equations are formulated over $s$ state measurements and the filter update step is,
\begin{align}\label{update_step}
    \boldsymbol{\Lambda}_{t+1|t+1} = \frac{\mathbf{M} \boldsymbol{\Lambda}_{t+1|t}}{1^T \mathbf{M} \boldsymbol{\Lambda}_{t+1|t}}
\end{align}
In Algorithm \ref{alg:filtering}, the inputs are measurements $\mathbf{y}_{t-s:t}$ and belief vector $\boldsymbol{\Lambda}_{t-s}$ over the past $s$ time steps. The output is the measurement likelihood $p(\hat{\mathbf{x}}_{t-s:t}=\mathbf{y}_{t-s:t} | \boldsymbol{\Lambda}_{t-s})$, where $\hat{\mathbf{x}}$ denotes the open-loop Nash solution of the dynamic game (\ref{gnep}). The game is solved over $H$ horizon from $\mathbf{y}_{t-s}$ with $\boldsymbol{\Lambda}_{t-s}$ and the control sequence $\mathbf{u}_{t-s:t+H}$ is computed. Then the dynamics is propagated forward up to time t and the state trajectory $\hat{\mathbf{x}}_{t-s:t}$ is computed. Finally the measurement likelihood $p(\hat{\mathbf{x}}_{t-s:t}=\mathbf{y}_{t-s:t} | \boldsymbol{\Lambda}_{t-s})$ is computed. 
 \begin{algorithm}
\caption{Measurement likelihood for discrete Bayesian filter}\label{alg:filtering}
\begin{algorithmic}[1]
\State \textbf{Input:} $\mathbf{y}_{t-s:t}, \boldsymbol{\Lambda}_{t-s}$
\State \textbf{Output:} $p(\hat{\mathbf{x}}_{t-s:t}=\mathbf{y}_{t-s:t} |\boldsymbol{\Lambda}_{t-s})$
\State \textbf{Steps:} $\mathbf{u}_{t-s:t+H} \leftarrow$ solve game (\ref{gnep}) from $\mathbf{y}_{t-s}$ with $\boldsymbol{\Lambda}_{t-s}$ over $H$ horizon
\State $\hat{\mathbf{x}}_{t-s:t} \leftarrow$ forward propagate (\ref{dyna}) up to time $t$
\State $p(\hat{\mathbf{x}}_{t-s:t}=\mathbf{y}_{t-s:t} | \boldsymbol{\Lambda}_{t-s}) \leftarrow $  likelihood with $p(\mathbf{x}_{t-s:t} | \boldsymbol{\Lambda}_{t-s}) \propto p(\mathbf{x}_{t-s:t} | \hat{\mathbf{x}}_{t-s:t}(\mathbf{u}_{t-s:t}(\boldsymbol{\Lambda}_{t-s})))$
\end{algorithmic}
\end{algorithm}

In our setup, at time step $t-s$ agent $i$ solves his current perceived game determined by $\boldsymbol{\Lambda}_t$ according to \eqref{kay}-\eqref{cjay} with horizon $H$ and executes the trajectory obtained for $s<H$ time steps. At step $t$, agent $i$ observes the executed joint state trajectories of all agents between $t-s$ and $t$ which we denote $\hat{X}_{t-s:t}= [\mathbf{\hat{x}}_{t-s}^{\top}, \dots, \mathbf{\hat{x}}_{t}^{\top}]^\top$. Similarly, for $I \in \{0,1,\dots,\xi\}$ agent $i$ can obtain hypothetical trajectories $X^I_{t-s:t}=[\mathbf{x}_{t-s}^{I,\top}, \dots, \mathbf{x}_{t}^{I,\top}]^\top$ by solving each of the available versions of the game, with the convention that $I=0$ represents the communicated game. We can thus compute, for hypotheses $I \in \{0,1,\dots,\xi\}$, a disparity score with the observed executed joint trajectory $d^I$, which we take to be the sum along all points of the trajectory if the $l1$-norm of the element-wise relative state error, such that,
\begin{align}
    d^I = \sum_{k=t-s}^{t} \norm{(\mathbf{x}_{k}^{I}-\mathbf{\hat{x}}_k)\oslash \mathbf{\hat{x}}_k}_1
    \label{rel_err}
\end{align}
where the $\oslash$ symbol represents the Hadamard element-wise division. The choice of this formulations allows for all elements of the observed state (positions, translation velocities, angular rates. etc.) to have the same weight in the norm computation. We shed light on the necessity of taking numerical precautions to avoid divisions by zero in \eqref{rel_err}. We also ensure all computed error scores are in practice strictly positive by adding a small offset to \eqref{rel_err}.
The hypothesis relative likelihoods $\lambda_I$ for $I \in \{0,1,\dots,\xi\}$ presented in Section \ref{filter} must rank in the inverse order of that of the disparity scores as closer trajectories are assumed to arise from games with more accurate cost function representations.  We obtain our current estimate of the relative likelihoods denoted $\tilde{\boldsymbol{\Lambda}}_{t} = [\tilde{\lambda}_0,\dots,\tilde{\lambda}_\xi]$ by requiring likelihoods to be inversely proportional to disparity scores and thus to satisfy the following conditions,
\begin{align}
    \tilde{\lambda}_I &= \frac{d^J}{d^I}\tilde{\lambda}_J, \quad \forall I,J \in \{0,1,\dots,\xi\}, \ I\neq J \label{inv_prop}\\
    \sum_{I=0}^ \xi \tilde{\lambda}_I &= 1 \label{sum21}\\
    \tilde{\lambda}_I &\geq 0, \quad \forall I\in \{0,1,\dots,\xi\} \label{ispos}
\end{align}
Note that the $\xi+1 \choose 2$ conditions ($\xi+1$ choose $2$) in \eqref{inv_prop} are highly redundant and can be replaced by $\xi$ conditions:
\begin{align}
    \tilde{\lambda}_I &= \frac{d^0}{d^I}\tilde{\lambda}_0, \quad \forall I \in \{1,\dots,\xi\} \label{inv_simp}
\end{align}
Also the strict positivity of $d^I$ for $I \in \{0,1,\dots,\xi\}$ combined with condition \eqref{sum21} ensures that any solution to \eqref{inv_simp} and \eqref{sum21} necessarily satisfies \eqref{ispos}. Thus, computing the current estimate of the relative likelihoods comes down to solving the simple linear system
\begin{align}\label{linear_system}
    \mathbf{A}_{t}\tilde{\boldsymbol{\Lambda}}_{t}^\top = \mathbf{B}
\end{align}
with $\mathbf{A}_{t} \in \mathbb{R}^{(\xi+1) \times (\xi+1)}$ and $\mathbf{B} \in \mathbb{R}^{\xi+1}$ such that,
\begin{align}
\mathbf{A}_{t} = 
\begin{bmatrix}
   1 & 1 & 1 & \cdots & 1 \\
   1 & -\frac{d^1}{d^0} & 0& \cdots & 0 \\
   1 & 0 & \ddots &  & \vdots \\
   \vdots  & \vdots &  & \ddots & 0  \\
   1 & 0 & \cdots& 0 & -\frac{d^\xi}{d^0} 
 \end{bmatrix}, \ 
 \mathbf{B}=
 \begin{bmatrix}
   1 \\
   0 \\
   \vdots \\
   \vdots  \\
   0.
 \end{bmatrix}
\end{align}
The likelihood estimation scheme is based on the online filtering approach presented in section \ref{filter}. For the filter predict step (\ref{predict_step}), the transition probability matrix is assumed as identity matrix $\mathbf{T} = \mathbf{I}_{\xi+1}$ to make minimal assumption on the transition dynamics between different hypothesis. Therefore $    \boldsymbol{\Lambda}_{t+1|t} =     \boldsymbol{\Lambda}_{t|t}$. Equation (\ref{linear_system}) is equivalent to the filter update step (\ref{update_step}). While the measurement likelihood $p(\hat{\mathbf{x}}_{t-s:t}=\mathbf{y}_{t-s:t} | \boldsymbol{\Lambda}_{t-s})$ discussed in algorithm \ref{alg:filtering} is computed using equation (\ref{rel_err}).
We finally obtain the new value of $\boldsymbol{\Lambda}_{t}$ used to define the next perceived game to solve via the fusion of the previous value $\boldsymbol{\Lambda}_{t-s}$ with the new estimate $\tilde{\boldsymbol{\Lambda}}_{t}$ according to a fixed update rate $\gamma$, such that,
\begin{align}
    \boldsymbol{\Lambda}_{t} = (1-\gamma) \boldsymbol{\Lambda}_{t-s} + \gamma \tilde{\boldsymbol{\Lambda}}_{t} \label{lam_update}
\end{align}

The update rate $\gamma$ is the filter's only hyperparameter. It determines how fast we update our likelihood vector according to previous estimates. Smaller $\gamma$ values (slower) can be useful in scenarios with high replanning frequencies, where shorter observed trajectories contain less useful signal, and a gradual convergence of $\boldsymbol{\Lambda}$ over a larger sample is desired. Larger $\gamma$ values favor quick adaptation to new observations contradicting the current active hypothesis.

\subsection{Proposed Motion Planner}
Our solution consists of an MPC implementation of the open-loop dynamic game solver over a time horizon $H$, with the Nash equilibrium policy subsequently executed for $H_N$ time steps before updating the perceived game costs and repeating.
Agent $i$, with cost function $c^i$, initial joint state of the scene $\mathbf{x}_0$, initial hypothesis likelihood vector $\boldsymbol{\Lambda}_0$, and hypothesis cost models for other agents $c_{h}^i$, thus computes its motion according to our decentralized on-board planner outlined in Algorithm \ref{alg:plan}.

\begin{algorithm}
\caption{Planner for robot $i$}\label{alg:plan}
\begin{algorithmic}[1]
\Procedure {$\textsc{Planner}(\mathbf{x}_0, \boldsymbol{\Lambda}_0, c^i, c_{h}^i, H, H_N, \gamma)$}{}
\State $\mathbf{x} \gets \mathbf{x}_0$, \quad $\boldsymbol{\Lambda} \gets \boldsymbol{\Lambda}_0$
    \While{$interacting$}{
        \State $\textsc{Communicate} (c^i)$
        \State $c_c^{\neg i} \gets \textsc{ReceiveCommunicatedCosts}$
        \State $c \gets \textsc{FormPerceivedGame}(\boldsymbol{\Lambda}, c_c^{\neg i}, c_{h}^i)$ \hfill\eqref{kay}-\eqref{cjay}
        \State $\pi \gets \textsc{SolveDynamicGame}(\mathbf{x},c,H)$ \hfill\eqref{gnep}
        \State $\mathbf{x}^i \gets \textsc{Execute}(\mathbf{x}^i, \pi^i, H_N)$
        \State $\mathbf{x}^{\neg i} \gets \textsc{ObserveState}(H_N)$
        \State $\boldsymbol{\Lambda} \gets \textsc{UpdateLikelihood}(\pi, \mathbf{x}, \gamma)$ \hfill\eqref{linear_system}-\eqref{lam_update}
    \EndWhile}
\EndProcedure
\end{algorithmic}
\end{algorithm}

The hypothesis likelihood update compute time is largely negligible relative to the dynamic game solver. Hence, the planning time scales linearly with the number of hypotheses as compared to using ALGAMES to plan a single version of the dynamic game. Indeed, for a single version of a 3 agent game, a 20 step look-ahead and a 5 step executed trajectory, the compute costs required for solving the game and updating the hypotheses likelihood vector are shown in Table \ref{tab:time}.
\begin{table}[h!]
\centering
\caption{Orders of magnitude of the run time, number of allocations and disk space required to solve the dynamic game and to update the hypothesis likelihood vector.}
\begin{tabular}{l c c c}
\toprule
\textbf{Task} & \textbf{Run time (s)} & \textbf{\# alloc.} & \textbf{Space (MB)} \\ 
\midrule
\textsc{SolveDynamicGame} & $10^{-1}$ & $10^5$ & 100 \\
\textsc{UpdateLikelihood} & $10^{-4}$ & $10^3$ & 0.1 \\
\bottomrule
\end{tabular}
\label{tab:time}
\end{table}


\section{Simulations}

We test our proposed solution on three autonomous driving scenarios inspired by real-life confusion cases with increasing complexity. We look at a two vehicle overtaking maneuver, a three vehicle lane merging scenario and a four vehicle intersection negotiation. In all the simulations, all agents plan their trajectories using our proposed planner.

\subsection{Simulations setup}

The dynamic games solver we use is ALGAMES \cite{LeCleac'h2019}, which handles trajectory optimization problems with multiple actors and general nonlinear state and input constraints. 
The vehicles obey nonlinear unicycle dynamics. The state of a vehicle $\mathbf{x}^i_t$ comprises of its 2D position, its heading angle and scalar velocity. The control input $\mathbf{u}^i_t$ comprises of the angular velocity and scalar acceleration.

\subsubsection{Constraints}
The dynamics constraints at time $t$ consist in following the system dynamics given by \eqref{dyna}, with $f$ being unicycle dynamics in all our driving simulations.
We also enforce collision-avoidance constraints on the trajectories, by modelling collision zones of the vehicles by circles or radius $r$, such that, at any time step $t$,
\begin{align}
    \norm{\mathbf{p}^i_t - \mathbf{p}^j_t}_2^2 \geq r^2, \quad \forall i,j \in \{1,\dots,N\}
\end{align}
In addition, we require the vehicles to remain on the road, by constraining the distance between each vehicle and the closest point $\mathbf{q}$ on each boundary $b$ to remain larger than the collision radius $r$,
\begin{align}
    \norm{\mathbf{p}^i_t - \mathbf{q}_b}_2^2 \geq r^2, \quad \forall b, \forall i \in \{1,\dots,N\}
\end{align}
where $ \mathbf{p}^i_t = [px_t^i, py_t^i]$ contains the plane coordinates of the agent $i$ at time $t$ extracted from the complete state vector $\mathbf{x}^i_t$.
Thus, the autonomous driving problem is formalized via non-convex and non-linear coupled constraints.

\subsubsection{Cost function}
The cost structure considered is quadratic, penalizing the distance to the desired final state $\mathbf{x}_f$ and the use of controls,
\begin{align}
    J^i(X,\pi^i) &= \sum_{t-1}^{H-1}\frac{1}{2}(\mathbf{x}_t-\mathbf{x}_f)^\top \mathbf{Q} (\mathbf{x}_t-\mathbf{x}_f) + \frac{1}{2} \mathbf{u}_t^{i,\top} \mathbf{R} \mathbf{u}_t^i \nonumber\\ & \hspace{1em} + \frac{1}{2}(\mathbf{x}_H-\mathbf{x}_f)^\top \mathbf{Q}_f (\mathbf{x}_H-\mathbf{x}_f) \label{eq:cost_LQR}
\end{align}
where $\mathbf{Q}$, $\mathbf{R}$ and $\mathbf{Q}_f$ represent state, input and final state penalization weight matrices, respectively.
This cost function depends only on the decision variables of vehicle $i$, as players’ behaviors are only coupled through collision constraints. Thus, although knowledge of other agents' intentions does not intervene in the individual cost agent $i$ is optimizing for, it does however determine the trajectories of others, and subsequently the collision constraints agent $i$ has to satisfy.


\subsection{Advantage of V2V communication}

\begin{figure}[!htb] \center \vspace{0.2cm} \epsfig{file=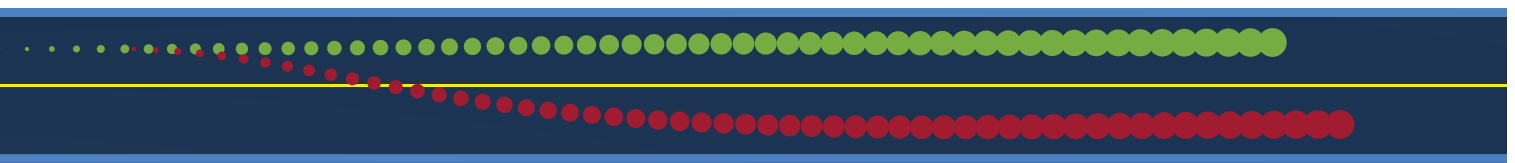, width=0.45\textwidth}

\vspace{1em}

\epsfig{file=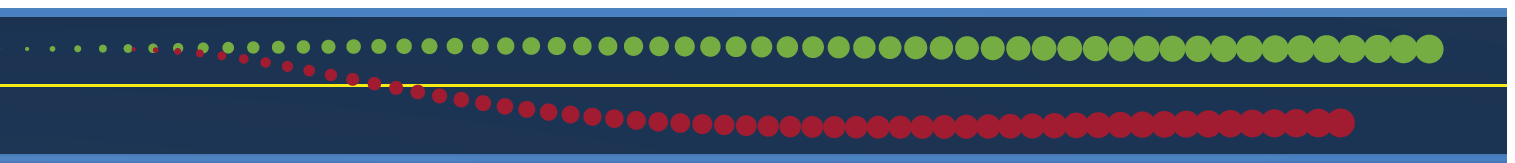, width=0.45\textwidth}
	\caption{No communication (top) and true cost communication (bottom). $v_1$ and $v_2$ executed trajectories in green and red respectively. Marker size increases with respect to time.} \vspace{-0.6cm}
	\label{fie:comm}
\end{figure}

We show through a simple takeover maneuver experiment that intention sharing between agents can improve the quality of the planned solutions. Vehicle $v_1$ wishes to maintain high velocity on the left lane along a two-lane highway. Vehicle $v_2$ is in front of it initially and is driving at a slower speed. $v_2$ wishes to move to the right lane. We consider two scenarios: first with $v_1$ agnostic to $v_2$'s intention, and second with the $v_2$'s true cost function available to $v_1$. In the first case, since no information is available to $v_1$, we let $v_1$ suppose that $v_2$ intends to continue driving on the left lane at the same speed. The trajectories in both cases are presented in Fig. \ref{fie:comm}.

We clearly notice that access to $v_2$'s intentions allows $v_1$ to satisfy its objective of maintaining high speed on the left. On the other hand, the absence of such information and of the ability to update the cost based on observations deteriorates the reward collected as $v_1$ gets stuck behind $v_2$, expecting the latter to drift back to the left lane.

\subsection{Unique true alternative}
We first validate the soundness of our approach by considering cases with one agent sending faulty communication. The other agents use our planner with one alternative hypothesis, which we here assume to be the true cost function of the faulty agent. Furthermore, we assume that all players initially trust each other (at $t=0$ we initialise $\lambda_0 = 1, \lambda_1=0$). 
We simulate three real-life driving situations: a takeover maneuver, a lane merging, and an intersection negotiation.

\subsubsection{Takeover}

Vehicle $v_1$ wishes to maintain high velocity on the left lane along a two-lane highway. In front of it on the same lane is a slower vehicle, $v_2$, which misleadingly expresses its intention to make way (e.g. by signalling to the right), when its true intention is to stay on the left lane.

\begin{figure}[!htb] \center \vspace{0.1cm} \epsfig{file=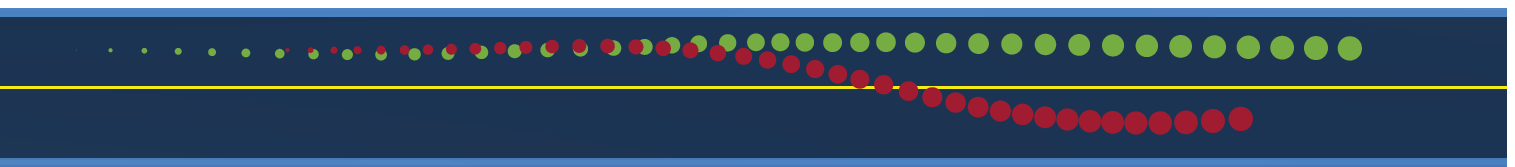, width=0.45\textwidth}

\vspace{1em}

\epsfig{file=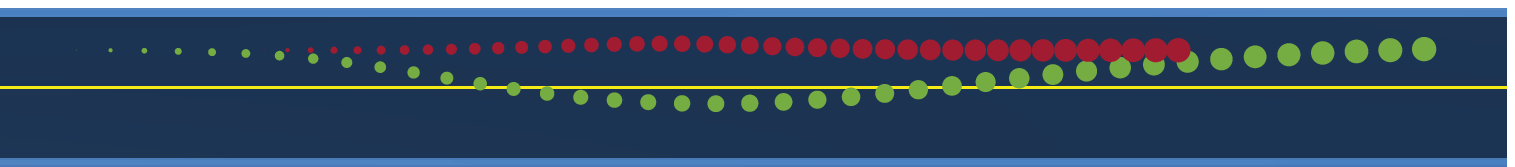, width=0.45\textwidth}
	\caption{Communication only planner solution (top) and adaptive hypothesis likelihood planner with $\gamma = 0.6$ solution (bottom). $v_1$ and $v_2$ executed trajectories in green and red respectively. Marker size increases with respect to time.} \vspace{-0.2cm}
	\label{fie:to_h1}
\end{figure}

The obtained trajectories with and without the use of our planner are presented in Fig. \ref{fie:to_h1}.

Relying on communicated cost functions only, the overtaking vehicle $v_1$ persists in attempting to pass by $v_2$ from the left as at each episode the Nash equilibrium with the fixed communicated cost from $v_2$ would yield an optimal policy in which $v_2$ moves out of $v_1$'s way. This leads to an aggressive maneuver where $v_1$ repeatedly bumps $v_2$ pushing it to the right lane for $v_1$ to squeeze through. Using our planner, $v_1$ observes $v_2$'s reluctance to make way over the first few iterations, and quickly assigns the alternative cost function to $v_2$, leading to a smooth overtake from the right.

\subsubsection{Lane-merging}

Next we consider a lane-merging maneuver involving three vehicles on a two-lane highway. Let $v_1$ be the source of the fault, communicating its willingness to make way to the merging vehicle $v_2$, when in truth it wishes to maintain driving along the right lane without reducing its speed. Driving along the left lane, $v_3$'s objective is to maintain high velocity.

\begin{figure}[!htb] \center \vspace{0.2cm} \epsfig{file=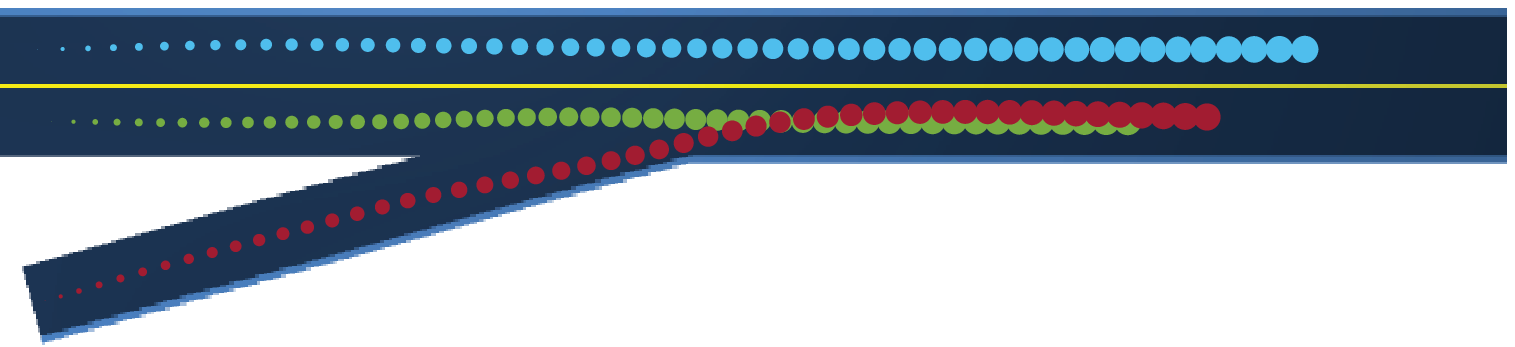, width=0.45\textwidth}

\vspace{1em}

\epsfig{file=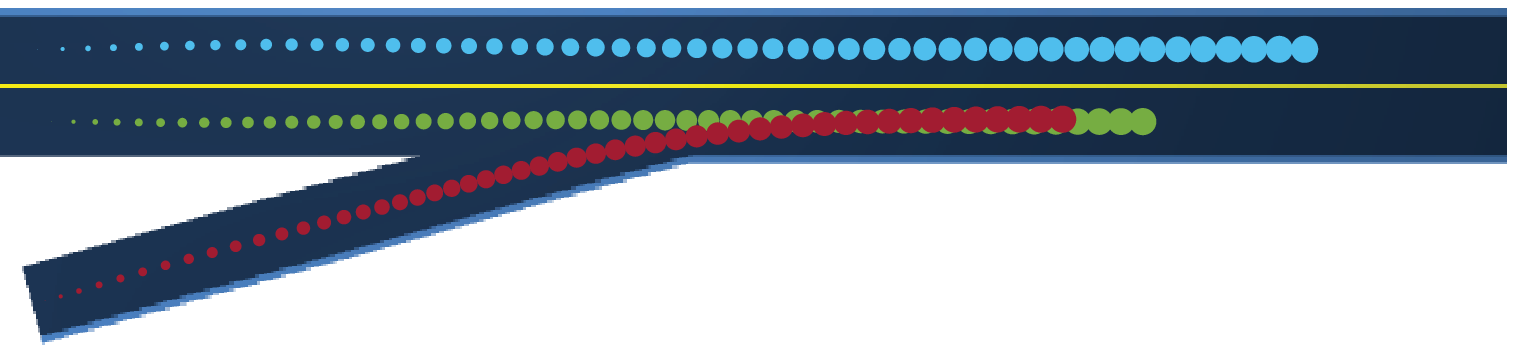, width=0.45\textwidth}
	\caption{Communication only planner solution (top) and adaptive hypothesis likelihood planner with $\gamma = 0.6$ (bottom) cases : $v_1$, $v_2$ and $v_3$ executed trajectories in green, red, and blue respectively. Marker size increases with respect to time.} \vspace{-0.2cm}
	\label{fie:3m_h1}
\end{figure}

Fig. \ref{fie:3m_h1} shows the trajectory obtained using communicated intentions only, and that with $v_2$ and $v_3$ using our planner.
The first trajectory depicts an aggressive and unsafe merging scenario. Indeed, at each planning step, $v_2$ persists in assuming $v_1$ will either slow down or switch to the left lane in order to accommodate it. Hence, $v_2$ continues to execute the merge in front of $v_1$, passing dangerously close to both $v_1$ and the road boundary. 
Using our adaptive planner, $v_2$, notices $v_1$'s reluctance to make way over a couple of planning episodes, assigns larger likelihood to the alternative scenario and subsequently plans to allow $v_1$ to pass first before merging. In this case, $v_2$ avoids finding itself in the hazardous situation of being tailgated by $v_1$. 


\subsubsection{4-way intersection}

\begin{figure}[!htb] \center \vspace{0.0cm} \epsfig{file=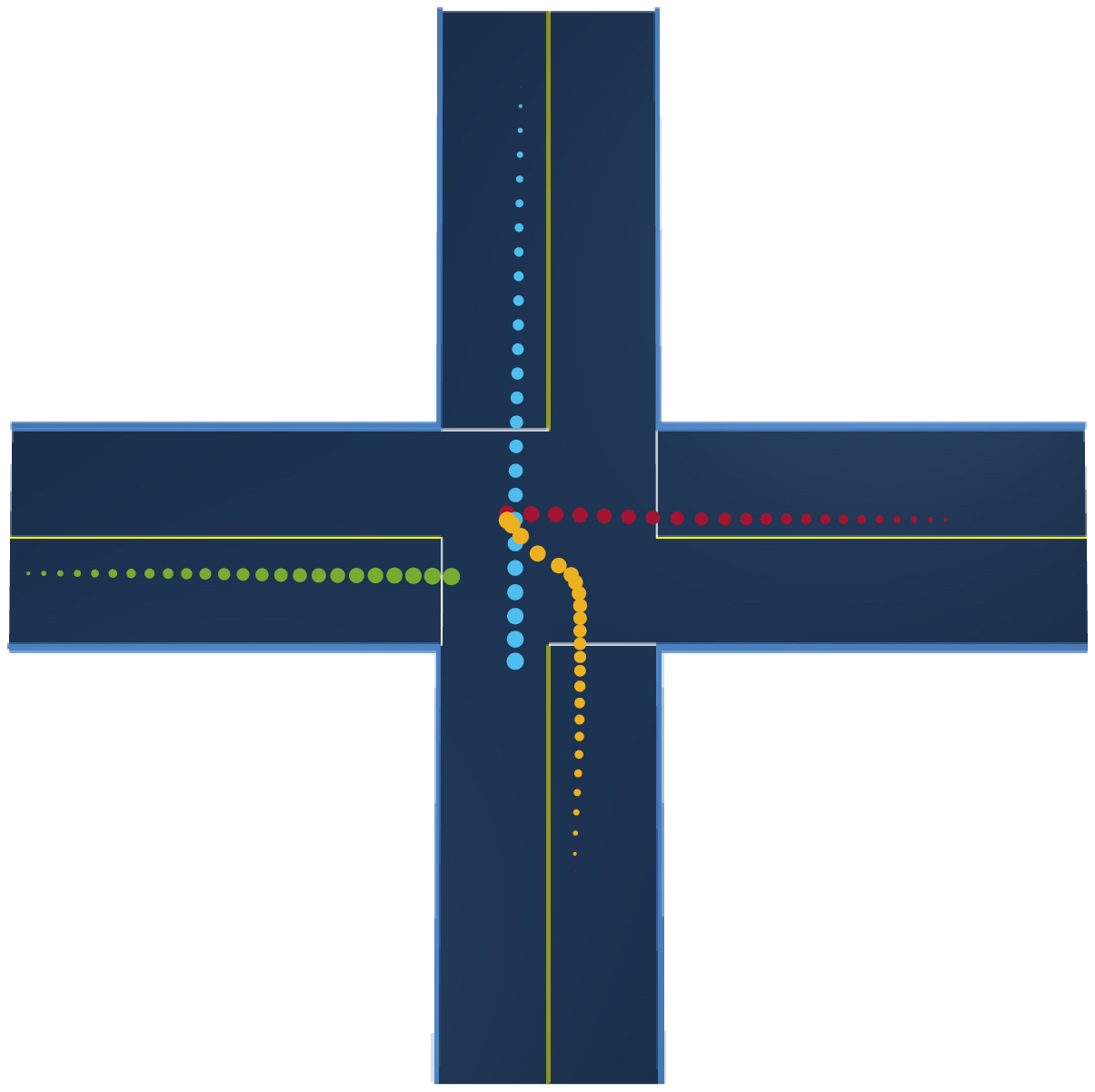, width=0.23\textwidth}
\epsfig{file=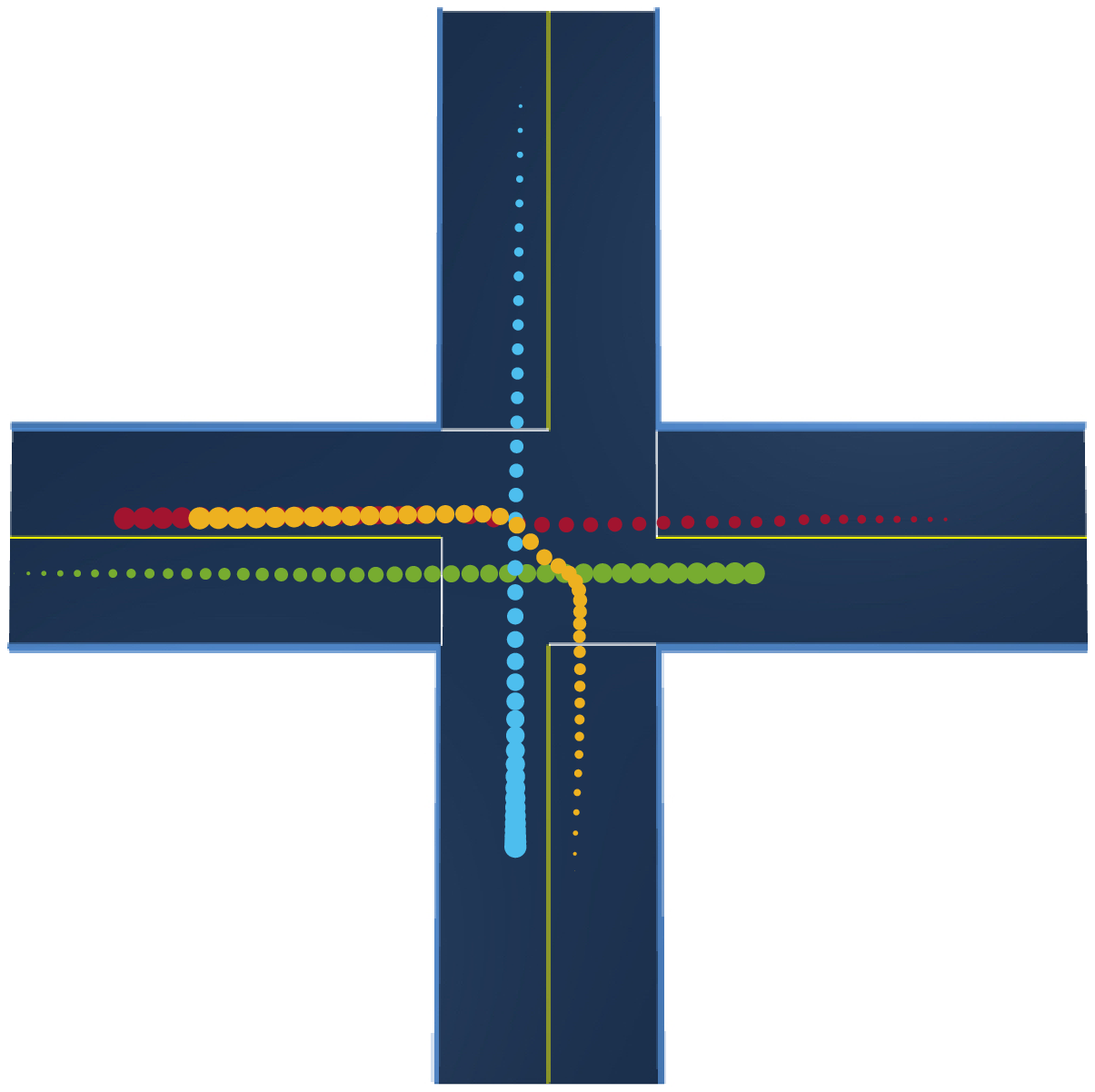, width=0.23\textwidth}
	\caption{Communication only planner solution (left) and adaptive hypothesis likelihood planner with $\gamma = 0.4$ (right) cases : $v_1$, $v_2$, $v_3$ and $v_4$ executed trajectories in green, red, blue and yellow respectively. Marker size increases with respect to time.} \vspace{-0.2cm}
	\label{fie:in_h1}
\end{figure}

The last example we consider involves four vehicles negotiating an intersection crossing. Three vehicles, $v_1$, $v_2$ and $v_3$ intend to drive straight across (west to east, east to west and north to south respectively). Vehicle $v_4$ wants to execute a left turn (south to west). 
We assume $v_4$ misleads the other agents about its aggressiveness. Trajectories planned with the faulty intention as well as those obtained with agents $v_1$, $v_2$ and $v_3$ using our adaptive planner are presented in Fig. \ref{fie:in_h1}.

Without adapting the cost functions, $v_2$ does not hasten its passage as $v_4$ comes increasing close to it, as communication from $v_4$ suggest it is willing to slow down to let $v_2$ through. Thus $v_2$ drives right in front of $v_4$ as it executes its turn, leading to a collision. In the adaptive case, $v_2$, noticing $v_4$'s absence of braking, accelerates as it enters the intersection and avoids getting crashed into.




\subsubsection{Update rate dependency}

\begin{figure}[!htb] \center
    \epsfig{file=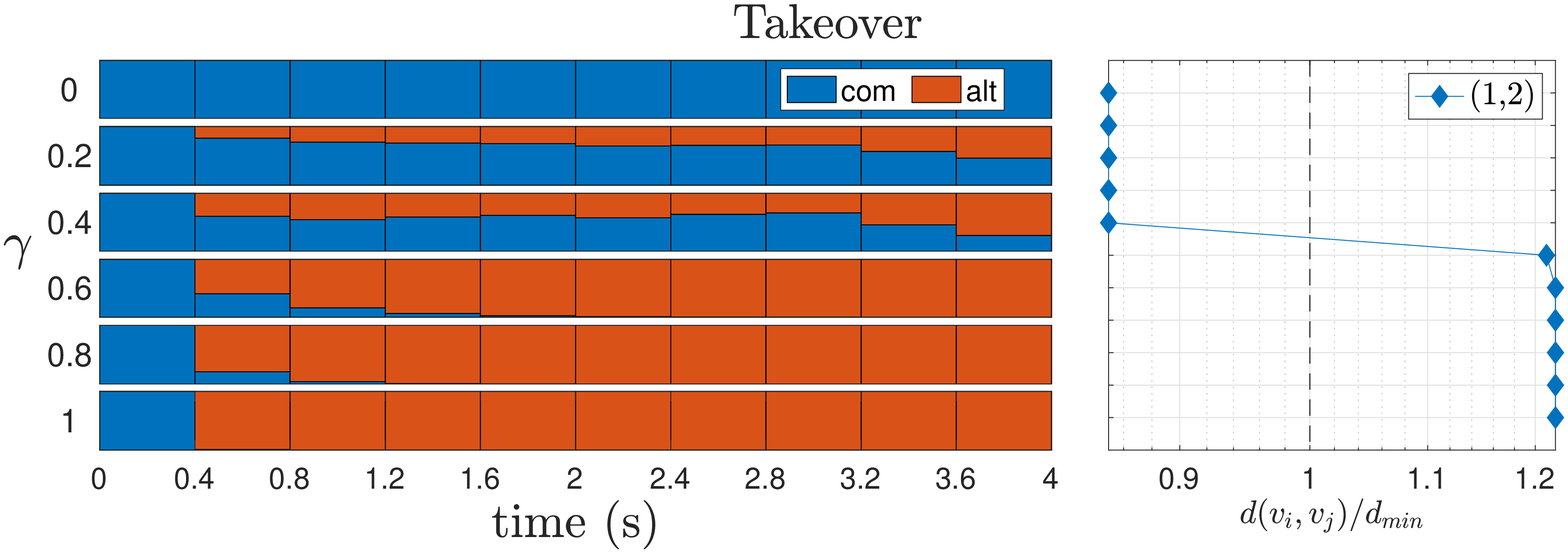, width=0.45\textwidth}
    \vspace{0.2cm}
    
    \epsfig{file=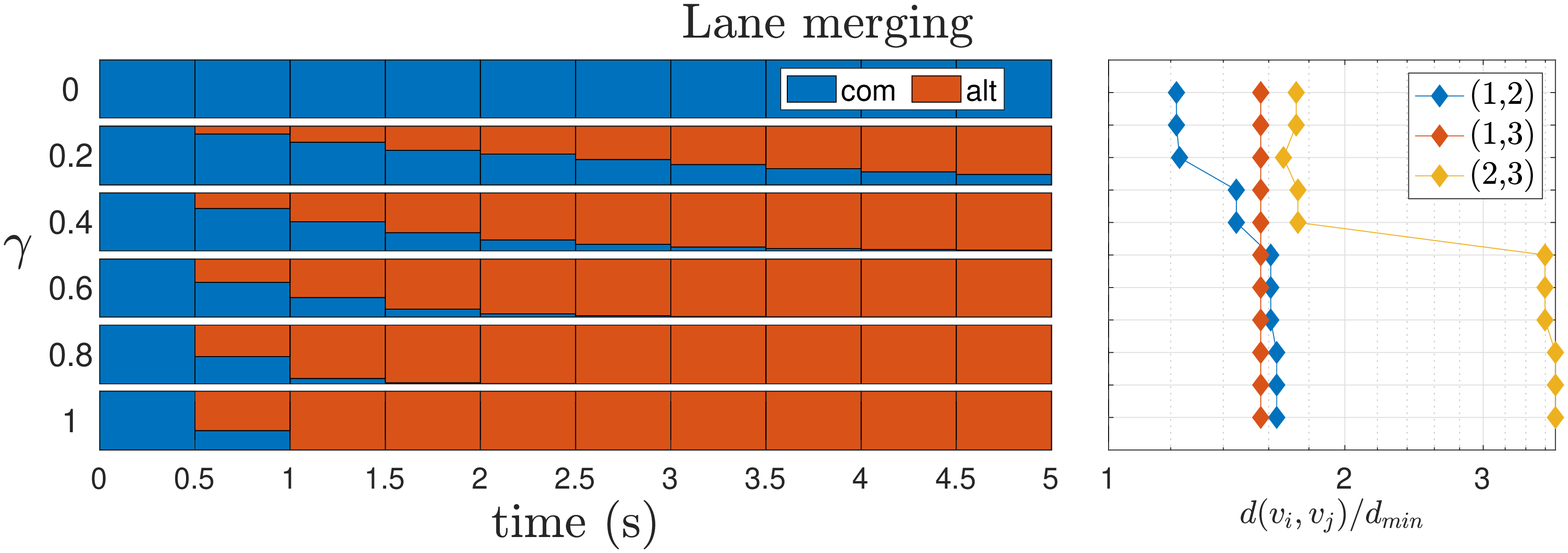, width=0.45\textwidth}
    \vspace{0.2cm}
    
    \epsfig{file=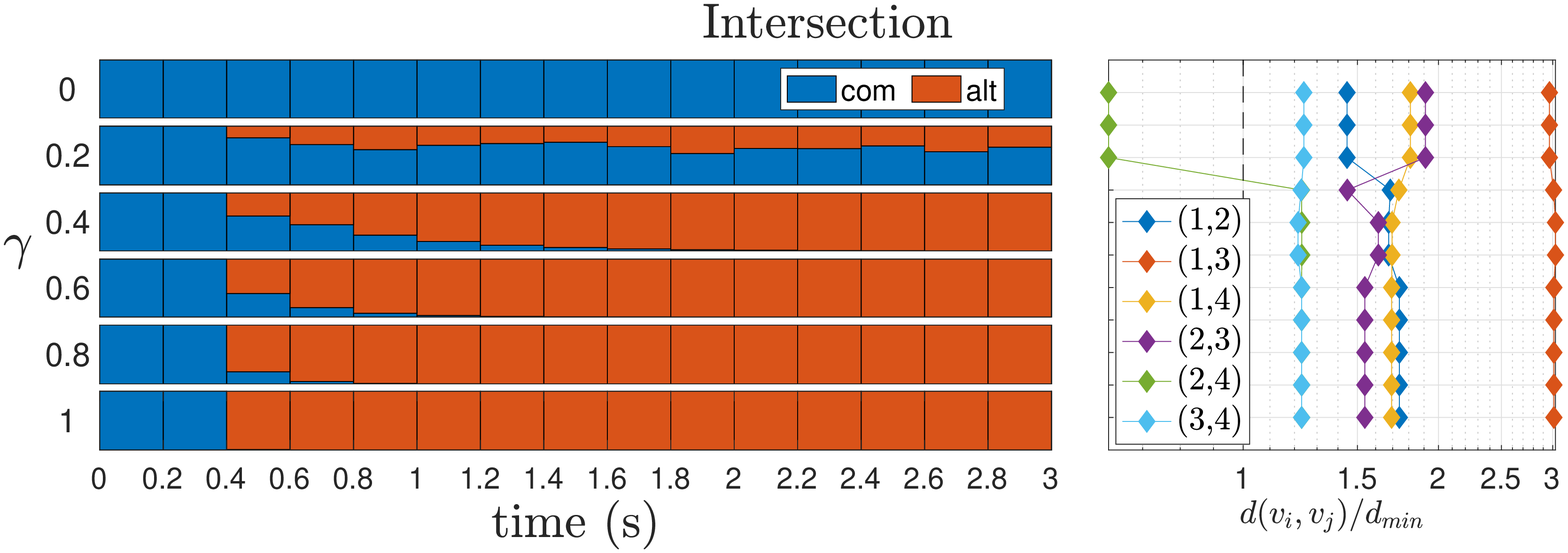, width=0.45\textwidth}
	\caption{Left: Evolution of the hypotheses relative likelihoods $\boldsymbol{\Lambda}$ throughout the simulation as a function of the filter update rate $\gamma$. The communicated cost's likelihood is depicted in blue, and the alternative hypothesis in red, the largest bar determining which version of the game is used for planning during the current episode. Right: Normalized distances between each pair of vehicles $d(v_i,v_j)/d_{min}$ (logarithmic scale) as a function of $\gamma$. The dashed line depicts the collision threshold $d_{min}$.} \vspace{-0.0cm}
	\label{fie:lnd}
	\vspace{-0.3cm}
\end{figure}

We study the effect of the update rate $\gamma$ on evolution of the perceived game determined by $\boldsymbol{\Lambda}$ and present quantitative analysis of the minimum distance between each pair of vehicles with respect to this filter parameter. Closed-loop simulations of our proposed planner for all three scenarios are presented in Fig. \ref{fie:lnd}. 

The main observation is that it is desirable for the hypothesis likelihood filter to converge before the start of the inter-vehicle interactions. Indeed, this not only allows agents to enjoy additional lead time to adapt their maneuvers to avoid collisions, but also to avoid interactions corrupting the intention signal used to update the filter. Overall, we notice that for large enough update rates, our planner allows agents to identify the correct intention of the faulty agent and use this information to replan safe trajectories.

\subsection{Multiple inexact alternatives}

\begin{figure}[!htb] \center
    \epsfig{file=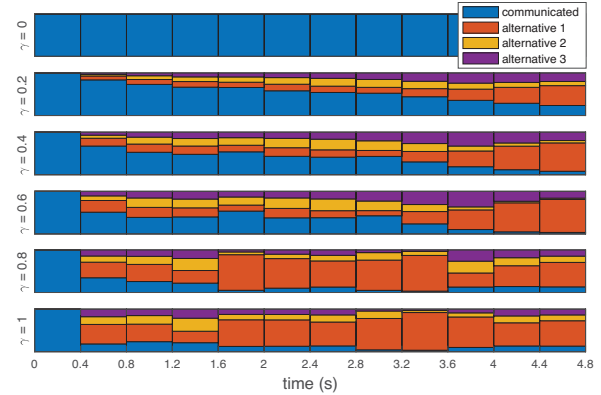, width=0.45\textwidth}
    %
    %
    \caption{Evolution of the hypotheses relative likelihoods $\boldsymbol{\Lambda}$ throughout the simulation as a function of the filter update rate $\gamma$. The communicated cost's likelihood is depicted in blue, and alternatives in red, yellow and purple. For $\gamma \geq 0.8$, collision is avoided.
    } \vspace{-0.2cm}
	\label{fie:tmulti_d}
\end{figure}

We next test and validate our planner in the more complex, realistic setting of multiple hypotheses about other agents' intentions, which have no reason to necessarily include the exact true cost functions. We show that when the hypothesis set contains a cost function that is close enough to the true one, safer planning is achieved using our approach. Indeed, we revisit the takeover example presented earlier, but now consider three alternative hypotheses to the communicated intention of $v_2$. The first is to maintain the left lane with a penalty on deviation from the lane center a quarter as small as the true one, the second is to maintain the left lane with a penalty 20 times smaller than the true value, and the third is to change lanes at a third of the initial speed. We show the evolution of $\boldsymbol{\Lambda}$ and of the minimum inter-vehicle distance as a function of $\gamma$ in Fig. \ref{fie:tmulti_d}.

According to the simulation results, our planner achieves collision-free trajectories in complex environments with broad and imprecise assumptions about other agents' costs utilizing the proposed online Bayesian filtering scheme which allows fast adaptation for the planner. The simulation results confirm that our solution can be used to ensure safety and improve performance in various driving scenarios.

\subsection{Performance and robustness}

We test the robustness of our approach to intention communication reliability, initial state conditions and process noise on the lane-merging scenario, comparing it to two planning baselines. We consider two available intention hypotheses for $v_1$ ($I_1$: $v_1$ wishes to stay on the right lane maintaining same speed, and $I_2$: $v_1$ wishes to accelerate with less emphasis on lane keeping), in addition to the communicated intention ($I_c$: $v_1$ is willing to yield the lane to the merging $v_2$). We assume all agents have access to the correct cost functions for $v_2$ and $v_3$. Our planner has access to all three hypotheses for planning. The first baseline is a communication-only planner using only $I_c$. The second uses our planner in a no-communication scheme, i.e. with only $I_1$ and $I_2$ as possible options. We uniformly randomize the initial positions of all three agents and introduce $20\%$ uniform process noise on the executed control commands. We run 100 merging simulations with faulty communication ($v_1$'s cost function is represented by $I_1$ and not $I_c$) and 100 runs with correct communication ($v_1$'s cost is $I_c$ as communicated). The results are presented in Table \ref{tab:results}.

\begin{table}[t]
\vspace{0.25cm}
\centering
\begin{tabular}{l c c c c c c }
\toprule
& \textbf{Hyp.} & $\boldsymbol{d}$ & \textbf{Risky} & \textbf{Crash} & $\boldsymbol{J}$ & $\boldsymbol{acc}_{max}$ \\ 
\midrule
\multirow{3}{*}{\rotatebox[origin=c]{90}{Faulty}}& $\{I_c\}$    & $1.40 \pm 0.09$ & 16\% & 0\% & 11.46 & 10.92\\
& $\{I_1, I_2\}$ & $1.56 \pm 0.04$ & 2\% & 0\%  & 11.21 & 10.80\\
& $\{I_c, I_1, I_2\}$  & $1.55 \pm 0.06$ & 1\% & 0\% & 11.07 & 10.74\\
\midrule
\multirow{3}{*}{\rotatebox[origin=c]{90}{Correct}}& $\{I_c\}$    & $1.57 \pm 0.01$ & 0\% & 0\% & 11.09 & 10.66\\
& $\{I_1, I_2\}$ & $1.54 \pm 0.12$ & 4\% & 2\% & 11.19 & 10.76\\
& $\{I_c, I_1, I_2\}$ & $1.57 \pm 0.01$ & 0\% & 0\% & 11.07 & 10.66\\  
\bottomrule
\multirow{3}{*}{\rotatebox[origin=c]{90}{\textbf{Total}}}& $\{I_c\}$ & $1.48 \pm 0.10$ & 8\% & 0\% & 11.28 & 10.79\\
& $\{I_1, I_2\}$ & $1.56 \pm 0.09$ & 3\% & 1\% & 11.20 & 10.78\\
& $\{I_c, I_1, I_2\}$  & $1.55 \pm 0.05$ & 0.5\% & 0\% & 11.07 & 10.70\\
\bottomrule
\end{tabular}
\caption{Statistical performance of our planner versus communication and agnostic baselines ($n=100$ manuevres for each of Faulty/Correct communication scenarios).
$d$ (mean $\pm$ std) represents the minimum normalized distance between each pair of vehicles, such that $d = \min\{d(v_i,v_j)/d_{min} \ | \ i,j \in \{1,2,3\}, \ i \neq j\}$, with $d_{min}$ denoting the collision threshold. Risky maneuvres are accounted for when $d$ is within $30\%$ of $d_{min}$, and crashes when $d\leq d_{min}$. The LQR cost of the maneuvre for the merging agent, denoted $J$, is computed according to equation \eqref{eq:cost_LQR}. The maximum acceleration of the merging vehicle $acc_{max}$ is used as a comfort score.}
\label{tab:results}
\vspace{-0.6cm}
\end{table}

As intended, our proposed planner generates safe trajectories whether the communicated intention is correct or not. Indeed, over the 200 simulations with our adaptive planner, only one trajectory is considered risky and no crashes are recorded. Also as expected, our planner matches the perfect information game in terms of performance when communication is correct. We also notice the importance correct communication can have in insuring safety, as not having access to a close enough intention in the set of hypotheses can lead to dangerous trajectories and crashes. Indeed, when $v_1$ correctly communicates its intention $I_c$ the adaptive planner using only $\{I_1, I_2\}$ exhibits a large variance in trajectories leading to a $2\%$ crash rate. 
Our planner is capable of both absorbing situations with faulty intention communication, and leveraging correct information for Nash optimal trajectory planning, without compromising on performance and comfort.

\section{Conclusion}
We have introduced a new receding horizon game-theoretic motion planner designed to operate in real-time in the framework of intention sharing multi-robot systems with potential communication faults. We demonstrate the robustness of our solution to such perturbations on complex autonomous driving scenarios, showcasing its ability to plan safe trajectories in situations from which naive implementations fail to recover. Future directions include integrating online intention identification into our framework in addition to incorporating state uncertainty.


\bibliography{main.bib}
\addtolength{\textheight}{-12cm}   




\end{document}